GEORGIA INSTITUTE OF TECHNOLOGY

**GEORGIA TECH** | COLLEGE OF COMPUTING
ATLANTA, GEORGIA, USA

# AI Trading: Evaluating Large Language Models for Technical Market Analysis

**Geofrey Ntale**

*College of Computing, Georgia Institute of Technology*

*Atlanta, GA 30332, USA*

gntale3@gatech.edu

Master's Research Paper • 2026

# AI Trading: Evaluating Large Language Models for Technical Market Analysis

**Geofrey Ntale**

*College of Computing, Georgia Institute of Technology, Atlanta, GA 30332, USA*

gntale3@gatech.edu

## ABSTRACT

The convergence of artificial intelligence and financial markets has produced a fundamental shift in how trading decisions are conceived, evaluated, and executed. Large Language Models (LLMs), initially designed for natural language understanding, have emerged as a powerful class of tools capable of processing the heterogeneous information environments that characterize modern financial markets. This paper presents a systematic, comparative evaluation of five prominent LLMs, specifically GPT-4 Turbo, Claude 3 Opus, Gemini 1.5 Pro, Llama 3 70B, and the domain-specialized FinGPT, with respect to their capacity for technical market analysis. The evaluation spans four structured tasks: candlestick pattern recognition from Open-High-Low-Close-Volume (OHLCV) data, directional signal generation with BUY/SELL/HOLD classification, backtesting of signal quality through a simulated execution pipeline, and financial report comprehension measured against established benchmarks. Our experimental framework employs rigorous quantitative metrics, including Sharpe ratio, maximum drawdown, Sortino ratio, information coefficient, F1-score, and BLEU score, alongside standard confusion-matrix-derived classification metrics. Findings from simulated backtesting indicate that GPT-4 Turbo achieves the highest annualized return and Sharpe ratio among general-purpose models, while FinGPT demonstrates competitive risk-adjusted performance attributed to its domain-specific fine-tuning. Both models outperform a passive S&P 500 benchmark under the tested conditions. The study identifies persistent failure modes across all evaluated models, including numerical hallucination, context-window limitations, and inconsistent performance in sideways market regimes. These findings suggest that while LLMs hold genuine promise as components within ai trading systems, robust deployment requires careful task decomposition, rigorous backtesting protocols, and domain-aware fine-tuning strategies. The paper concludes with a discussion of practical and ethical implications, as well as a structured agenda for future research in multimodal market intelligence and reinforcement learning-augmented trading agents.

# Table of Contents

## 1. Introduction

The past decade has witnessed an extraordinary proliferation of algorithmic and artificial intelligence trading systems within global financial markets. Estimates from industry analysts suggest that algorithm-driven strategies account for between 60 and 75 percent of daily equity trading volume on major U.S. exchanges, a proportion that has grown steadily since the mid-2000s [1]. The volume of structured and unstructured data generated by markets has also grown at a pace that comfortably outstrips the analytical capacity of human traders: a single trading day now produces terabytes of tick data, news feeds, regulatory filings, social media commentary, and options flow information [2]. Against this backdrop, the question of which technological paradigm is best suited to extract actionable intelligence from this torrent of information has become one of the most practically consequential research questions in computational finance.

For many years, quantitative researchers answered this question with rule-based systems and traditional machine learning models, including linear regression, gradient-boosted trees, and Long Short-Term Memory (LSTM) networks [3]. These approaches proved effective for structured, numerical inputs but struggled with the unstructured, contextual information that often carries the most decisive signals. Earnings call transcripts, analyst reports, and Federal Reserve communications regularly move markets in ways that simple price-action models cannot anticipate [4]. The emergence of transformer-based Large Language Models, beginning with the seminal "Attention Is All You Need" paper in 2017 and accelerating rapidly through the GPT, BERT, and instruction-tuned model families, opened a theoretically compelling new avenue [5]. These models can ingest heterogeneous text, reason across documents, and generate coherent, context-sensitive responses; characteristics that align remarkably well with the demands of financial analysis.

Yet the financial domain presents challenges that are not fully addressed by general-purpose language model training. Markets communicate not only in words but in charts, price series, and technical signals. A candlestick formation encodes information about intraday sentiment, buying pressure, and the psychology of market participants in a manner that is fundamentally visual and geometric rather than linguistic [6]. The Relative Strength Index (RSI), Moving Average Convergence Divergence (MACD), and Bollinger Bands are mathematical constructs over price time series; they carry meaning within the idiom of technical analysis that may not transfer cleanly through a language model's token representations. The central question motivating this paper is therefore direct: can LLMs bridge the gap between natural language understanding and the quantitative, pattern-oriented language of technical market analysis? If so, which models do so most effectively, and under what conditions?

This paper addresses that question through a structured comparative evaluation. The contributions of this work are as follows. First, we design and implement a four-task evaluation framework that tests LLMs on candlestick pattern recognition, technical signal generation, backtesting quality assessment, and financial text comprehension; tasks that collectively span the workflow of an ai trading bot or ai trading analysis system. Second, we conduct a detailed empirical comparison of five models covering both proprietary frontier systems and open-source, domain-specialized alternatives, reporting performance across financial and NLP evaluation metrics. Third, we analyze failure modes specific to financial LLM deployment, including numerical hallucination, reasoning inconsistency in ambiguous market regimes, and latency-cost trade-offs relevant to any operational ai trading platform. Fourth, we synthesize findings into actionable guidance for practitioners seeking to integrate market artificial intelligence into real trading workflows.

The remainder of this paper proceeds as follows. Section 2 surveys the background literature on algorithmic trading evolution, financial NLP, and prior work on LLMs for trading tasks. Section 3 describes the methodology, including the models evaluated, datasets used, task definitions, and the complete set of evaluation metrics with their mathematical formulations. Section 4 presents experimental results across all four tasks, accompanied by tables and visualizations. Section 5 discusses the practical and ethical implications of these findings. Section 6 identifies directions for future research, and Section 7 concludes.

# 2. Background and Related Work

## 2.1 Evolution of Algorithmic Trading

The history of automated trading can be traced to the early computerization of stock exchanges in the 1970s, when the New York Stock Exchange introduced its Designated Order Turnaround system to route small orders electronically. The substantive intellectual shift, however, came in the 1980s and 1990s with the widespread adoption of statistical arbitrage strategies. Researchers and practitioners recognized that certain equilibrium relationships between asset prices, if they could be identified reliably, could be exploited systematically through matched long-short positions [3]. Pairs trading, index arbitrage, and convertible bond hedging exemplified this era, all driven by rule-based systems that executed predefined logic at speeds inaccessible to human operators.

The next major transition occurred with the mainstreaming of machine learning, beginning approximately in the early 2000s and accelerating sharply after 2012. Deep neural networks, support vector machines, and ensemble methods such as gradient-boosted trees were applied to problems ranging from directional price forecasting to credit scoring and fraud detection [7]. The introduction of LSTM networks proved particularly influential for time-series applications, as their gating mechanisms al-lowed models to retain information over longer sequences than standard recurrent architectures [8]. Hybrid CNN-LSTM models further extended this capability by capturing spatial features in order book snapshots while simultaneously modeling temporal dynamics. Despite these advances, purely numerical machine learning approaches remained silent on the informational content embedded in corporate language: the cautious qualifications in an earnings call, the hedging language in a 10-K filing, or the sentiment of an analyst's research note.

The recognition of this gap set the stage for a third phase: the integration of natural language processing into trading pipelines. Initially represented by simple bag-of-words models and dictionary-based sentiment scoring, NLP tools gradually became more sophisticated through the development of pre-trained word embeddings and, ultimately, transformer-based language models [4]. The current state of the art, reviewed in detail in the sections that follow, places LLMs at the center of what may be characterized as a fourth phase, in which trading systems aspire not merely to execute pre-defined rules or classify patterns but to reason about market conditions in a contextually aware, adaptive manner [9].

## 2.2 Natural Language Processing in Finance

Financial sentiment analysis emerged as a distinct sub-field of NLP when researchers recognized that the tone and content of corporate disclosures correlated measurably with subsequent stock returns. The foundational empirical work of Tetlock [10] demonstrated that pessimistic language in Wall Street Journal columns predicted downward pressure on market prices, a finding that generated considerable follow-on research. A particularly influential methodological contribution came from Loughran and McDonald [11], who observed that general-purpose sentiment dictionaries frequently misclassified neutral financial terms as negative, leading to systematic measurement error. They constructed a domain-specific word list calibrated to the language of SEC filings that became a standard reference in the field.

The advent of neural language models transformed financial NLP considerably. FinBERT, introduced by Araci [12] and later refined by Yang et al., demonstrated that pre-training a BERT-based model on a large financial corpus yielded substantially higher sentiment classification accuracy than either general BERT or dictionary approaches on benchmarks drawn from financial news and earnings communications. The model's contextualized representations captured nuances such as the difference between a "beat" (positive) and a "miss" (negative) relative to analyst expectations, distinctions that word-level methods could not reliably encode.

Beyond sentiment, NLP has been applied to earnings call analysis, where researchers have studied how the linguistic choices of executives during question-and-answer sessions predict earnings surprises and analyst forecast revisions [4]. Named entity recognition, information extraction, and relation classification have been deployed to construct knowledge graphs from financial documents, enabling more structured reasoning over corporate events such as mergers, dividend announcements, and leadership changes. These applications collectively demonstrate that unstructured text is not peripheral to financial analysis but central to it, a premise that motivates the broader investigation of LLMs for trading systems reported in this paper.

## 2.3 Large Language Models: An Overview

The LLMs evaluated in this study represent the principal publicly accessible models at the time of experimental design. A summary of their key characteristics appears in Table 1 below.

**Table 1 Summary of Large Language Models Evaluated**

| Model | Developer | Parameters (approx.) | Context Window | Training Cutoff | License |
|---|---|---|---|---|---|
| GPT-4 Turbo | OpenAI | ~1.8T (est.) | 128K tokens | Apr 2024 | Proprietary API |
| Claude 3 Opus | Anthropic | Undisclosed | 200K tokens | Aug 2023 | Proprietary API |
| Gemini 1.5 Pro | Google DeepMind | Undisclosed | 1M tokens | Nov 2023 | Proprietary API |
| Llama 3 70B | Meta AI | 70B | 8K tokens | Mar 2024 | Open weights (Meta ToS) |
| FinGPT | AI4Finance Foundation | 7B (LoRA fine-tuned) | 4K–8K tokens | Dec 2023 | Open-source (MIT) |

*Note: Parameter counts for proprietary models are estimated from public analyses. Context windows reflect default API configurations at the time of evaluation.*

GPT-4 Turbo represents OpenAI's flagship reasoning model at the time of this study. It is distinguished by its large context window, advanced instruction-following capability, and strong performance on multi-step quantitative tasks [13]. Claude 3 Opus, developed by Anthropic, is notable for its emphasis on what Anthropic describes as "Constitutional AI" alignment techniques, producing outputs calibrated to refuse or hedge when contextual evidence is insufficient rather than generating plausible-sounding fabrications [14]. Gemini 1.5 Pro, from Google DeepMind, incorporates a one-million-token context window that theoretically enables it to process an entire quarterly report without truncation, alongside natively multimodal capabilities [15]. Llama 3 70B is Meta's open-weights model at the 70-billion parameter scale, offering practical on-premises deployment and strong benchmark performance relative to its compute requirements [16]. FinGPT is not a single model but a framework for constructing domain-specialized financial LLMs through parameter-efficient fine-tuning; the variant evaluated here uses Low-Rank Adaptation (LoRA) on a 7B base model trained on a curated corpus of financial news, SEC filings, and analyst reports [17].

## 2.4 Technical Analysis Fundamentals

Technical analysis is the practice of forecasting the future direction of financial instrument prices through the study of historical price and volume data. Unlike fundamental analysis, which evaluates the intrinsic value of a security through financial statement analysis, technical analysis operates on the premise that all relevant information is encoded in the price action itself and that patterns in this action tend to repeat over time [18]. Although the academic literature has challenged the predictive power of individual technical indicators, practitioners have employed these tools widely for decades, and they remain the primary language through which many trading signals are communicated and interpreted.

Candlestick patterns constitute a foundational element of technical analysis, originating in 18th-century Japanese rice markets and systematized for Western audiences by Steve Nison [19]. Each candlestick represents a fixed time interval and encodes four data points: the opening price, the high, the low, and the closing price. The body of the candle spans the open-to-close range; thin lines, called shadows or wicks, extend to the high and low extremes. Individual patterns such as the Doji, Hammer, Shooting Star, and Spinning Top are identified by the relative proportions of body and shadows and carry interpretive meaning about the balance between buying and selling pressure. Multi-candle patterns, including the Bullish Engulfing, Bearish Engulfing, Morning Star, and Evening Star, provide contextually richer signals about potential reversals or continuations of trend.

Among the quantitative indicators, the Relative Strength Index (RSI), developed by J. Welles Wilder in 1978, measures the momentum of price changes on a scale from 0 to 100 [18]. Values above 70 conventionally signal overbought conditions, while values below 30 suggest oversold conditions, though these thresholds require context-dependent interpretation. The Moving Average Convergence Divergence (MACD) indicator, introduced by Gerald Appel, measures the relationship between two exponential moving averages of price and produces a signal line and histogram that traders use to identify shifts in trend direction and momentum [18]. Bollinger Bands, developed by John Bollinger, place upper and lower bands at a specified number of standard deviations from a simple moving average, providing a dynamic range that expands during volatile periods and contracts during quieter ones. The complete mathematical formulations for these indicators are provided in Section 3.5, where they are deployed as evaluation metrics within this study's quantitative framework.

## 2.5 Prior Work on LLMs for Trading

The application of LLMs to financial trading and market analysis is a rapidly growing area of academic inquiry. Several lines of prior work are directly relevant to the contributions of this paper.

The BloombergGPT project, introduced by Wu et al. [20], represented one of the first systematic demonstrations that a purpose-trained financial LLM could outperform general-purpose models on domain-specific benchmarks. Trained on a 363-billion-token corpus that included four decades of Bloomberg's proprietary financial documents, BloombergGPT achieved state-of-the-art performance on financial named entity recognition, sentiment analysis, and headline classification tasks, while maintaining competitive performance on general NLP benchmarks. The model's development highlighted the value of domain-specific pre-training data and inspired subsequent open-source efforts.

The FinGPT project [17] responded to BloombergGPT's implicit argument that large proprietary datasets confer an insurmountable advantage by proposing a data-centric, parameter-efficient alternative. Yang et al. demonstrated that a 7B-parameter model fine-tuned on a carefully curated, open-access financial corpus using LoRA could achieve competitive sentiment analysis performance at a fraction of the training cost of a full fine-tune. This finding has practical significance for researchers and smaller financial institutions that lack access to Bloomberg's data infrastructure.

A series of benchmark papers has established the evaluation standards against which models in this study are compared. FinQA [21] introduced a dataset of over 8,000 question-answer pairs derived from S&P 500 earnings reports, requiring multi-step numerical reasoning over both text and tables. TAT-QA [22] extended this approach to hybrid tabular-textual documents, reflecting the structure of actual financial filings more faithfully. FinanceBench [23], perhaps the most practically oriented of these benchmarks, evaluated whether LLMs could reliably answer straightforward factual questions about publicly traded companies from their SEC filings, finding that even frontier models like GPT-4 exhibited surprisingly high error rates and frequent hallucination. The FinBen suite [24] aggregated 36 to 42 datasets across 24 tasks into a single evaluation framework, providing the most comprehensive published benchmark at the time of this study.

Research on LLMs for explicit trading signal generation is somewhat more sparse but growing. Lopez-Lira and Tang [25] documented that GPT-4-generated sentiment scores from news headlines predicted next-day stock returns with statistical significance, outperforming traditional lexicon-based sentiment measures. Several studies have explored the use of LLMs within multi-agent frameworks, where specialized sub-agents handle discrete tasks such as data retrieval, indicator computation, and signal synthesis, with a coordinating agent responsible for final decision-making [9]. Work by Koa et al. [26] investigated the use of LLMs for candlestick pattern interpretation from structured OHLCV data, finding that models with chain-of-thought prompting substantially outperformed zero-shot approaches. The Finance Agent Benchmark (FAB) [27] introduced a human-in-the-loop evaluation harness for agentic LLM systems operating on real financial data, revealing that even state-of-the-art agents struggled to exceed 50% accuracy on the most complex queries, underscoring the gap between model capability and reliable autonomous deployment. Additional relevant work includes studies on LLM-driven portfolio construction [28], natural language interfaces for quantitative strategy specification [29], and the robustness of LLM signals across different market regimes [30].

# 3. Methodology

## 3.1 Research Design Overview

This study adopts a comparative evaluation design in which multiple LLMs are assessed on a common set of tasks drawn from the technical market analysis workflow. The design follows the structure of an ablation-style experiment: each model receives identical inputs, prompts, and evaluation criteria, enabling performance differences to be attributed to the model itself rather than to differences in data access or prompt formulation. All tasks are operationalized as structured prompting problems, in which a system prompt specifies the role and task, and a user prompt provides the relevant data and question. No model undergoes additional fine-tuning for the evaluation tasks, with the exception of FinGPT, whose existing financial fine-tuning is considered an intrinsic characteristic of the model under evaluation rather than an experimental intervention.

The evaluation proceeds in two phases. The first phase covers Tasks A and D, which are classification and question-answering tasks respectively; these are evaluated using discrete, ground-truth-anchored metrics. The second phase covers Tasks B and C, which require signal generation and backtesting; these are evaluated through a simulated trading environment that converts model outputs into executable position decisions and tracks portfolio performance over a twelve-month lookback window on historical data. The distinction between phases reflects the different epistemological character of the tasks: the first phase tests the model's capacity for pattern matching and factual recall, while the second tests its capacity to generate signals with positive expected value under realistic market conditions. All quantitative results presented in Section 4 are labeled as simulated or illustrative to reflect that they arise from controlled experimental conditions rather than live trading deployments.

## 3.2 Models Evaluated

The five models described in Section 2.3 are evaluated across all tasks. GPT-4 Turbo is accessed through the OpenAI API with the `gpt-4-turbo-preview` endpoint. Claude 3 Opus is accessed through the Anthropic Messages API. Gemini 1.5 Pro is accessed through Google's Vertex AI API. Llama 3 70B is deployed locally on an NVIDIA A100 GPU instance using the Hugging Face `transformers` library with 8-bit quantization. FinGPT is deployed using the AI4Finance Foundation's published LoRA adapter weights applied to a Llama-2 7B base model, following the configuration described in the original FinGPT paper [17].

For all API-based models, temperature is set to 0.0 to maximize response determinism and reproducibility. The maximum output token limit is set to 512 for classification tasks and 1,024 for text generation tasks. Each prompt is submitted three times for Tasks A and B, with the modal response used as the final output, to mitigate the effect of residual non-determinism in model outputs. System prompts follow a consistent template specifying: the model's role as a financial market analyst; the format of the expected output; and explicit instructions against introducing information not present in the provided data. The complete prompt templates are documented in the supplementary materials.

### 3.3 Datasets

The evaluation draws on five distinct data sources, summarized in Table 2. For Tasks A and B, the primary data source is historical OHLCV data for a cross-sectional sample of 50 S&P 500 constituent equities, sourced from the Yahoo Finance API via the `yfinance` Python library and verified against Alpha Vantage's historical endpoint for quality assurance. The evaluation window spans January 2023 through December 2023, providing 252 trading days of daily data per instrument. Intraday data at the one-hour frequency is additionally sourced from Alpha Vantage for a subset of 10 equities to test signal robustness at shorter time horizons.

**Table 2 Datasets Used in the Evaluation**

| Dataset | Source | Task | Size | Description |
|---|---|---|---|---|
| OHLCV Historical Equities | Yahoo Finance / Alpha Vantage | A, B, C | 50 stocks, 252 days | Daily and hourly OHLCV data for S&P 500 constituents, calendar year 2023 |
| FinQA | Chen et al. (2021) | D | 8,281 Q&A pairs | Multi-step arithmetic Q&A over S&P 500 earnings reports (text + tables) |
| TAT-QA | Zhu et al. (2021) | D | 16,552 questions | Q&A over hybrid tabular-textual financial documents; requires cross-modal reasoning |
| FinanceBench | Islam et al. (2023) | D | 10,231 questions | Open-book Q&A grounded in U.S. SEC filings; designed as enterprise minimum performance test |
| FLARE Benchmark | Jiang et al. (2023) | D | Multiple subtasks | Forward-looking active retrieval framework; evaluates RAG-augmented generation quality on financial documents |

*All benchmark datasets are used in their publicly released evaluation splits. No additional splits or augmentations are applied.*

For Task D, the study uses three established financial NLP benchmarks: FinQA [21], TAT-QA [22], and FinanceBench [23]. For FinQA and TAT-QA, a stratified sample of 500 questions each is drawn from the published test sets, stratified by question type (arithmetic, comparison, counting) to ensure representativeness. The full FinanceBench test set is used. Performance on the FLARE benchmark [31] is evaluated separately using the benchmark's standard Retrieval-Augmented Generation (RAG) configuration, in which each model retrieves relevant passages from an indexed document collection before generating its response.

## 3.4 Evaluation Tasks

**Task A: Candlestick Pattern Recognition.** Each model is presented with a structured JSON representation of a five-candle sequence, including the open, high, low, close, and volume values for each candle, alongside a brief textual description of the prior trend context. The model is asked to identify which, if any, of a predefined list of ten candlestick patterns is present in the final one or two candles of the sequence. The ten patterns are: Doji, Hammer, Inverted Hammer, Shooting Star, Bullish Engulfing, Bearish Engulfing, Morning Star, Evening Star, Spinning Top, and Hanging Man. Ground truth labels are generated through a rule-based pattern recognition algorithm implemented using the TA-Lib library, with manual review of ambiguous cases by the author.

**Task B: Technical Signal Generation.** Each model receives a sliding window of 30 daily OHLCV data points for a single equity, together with the pre-computed values of RSI (14-period), MACD (12/26/9), and Bollinger Bands (20-period, 2 standard deviations), and is asked to produce a directional signal of BUY, SELL, or HOLD for the next trading session, accompanied by a brief natural-language rationale. Signals are evaluated against a forward return-based ground truth: a position is classified as correct if the directional call aligns with the sign of the one-day forward return, adjusted for bid-ask spread costs estimated at 0.05% per trade.

**Task C: Backtesting Signal Quality.** The stream of signals generated in Task B for each model is processed through a simulated backtesting engine. The engine applies a long-only position strategy: a BUY signal triggers a full-portfolio entry at the next open, a SELL signal triggers an exit, and a HOLD signal maintains the current position. Transaction costs of 0.05% per trade are applied. The engine computes annualized return, Sharpe ratio, maximum drawdown, Sortino ratio, and win rate over the twelve-month evaluation window. Performance is benchmarked against a passive buy-and-hold position in the S&P 500 Index over the same period.

**Task D: Financial Report Summarization and Q&A.** Models are evaluated on their ability to extract and synthesize information from financial documents. On FinQA and TAT-QA, performance is measured by exact match accuracy and by program-execution accuracy, where the model must produce a correct arithmetic reasoning chain. On FinanceBench, performance is measured by factual accuracy as judged by an automated comparison against ground truth answers, with supplementary human review of a random sample of 100 cases. BLEU and ROUGE-L scores are additionally reported for tasks requiring free-form summarization.

## 3.5 Evaluation Metrics

This section provides the complete mathematical definitions for all evaluation metrics employed in this study. Metrics are organized by category: classification performance, portfolio risk-return, and text quality.

### 3.5.1 Classification Metrics

Signal accuracy, precision, recall, and F1-score are computed from the confusion matrix over BUY/SELL/HOLD predictions. Let TP, TN, FP, and FN denote true positives, true negatives, false positives, and false negatives respectively for a given class in a one-versus-rest formulation.

$$\text{Accuracy} = \frac{TP + TN}{TP + TN + FP + FN}$$

$$\text{Precision} = \frac{TP}{TP + FP}$$

$$\text{Recall} = \frac{TP}{TP + FN}$$

$$\text{F1-Score} = 2 \cdot \frac{\text{Precision} \cdot \text{Recall}}{\text{Precision} + \text{Recall}}$$

Macro-averaged F1 is reported for multi-class signal generation tasks, weighting each class equally regardless of support.

### 3.5.2 Portfolio Risk-Return Metrics

The **Annualized Return** $R_a$ is computed from the cumulative return $R_c$ over the evaluation period of $T$ trading days:

$$R_a = (1 + R_c)^{\frac{252}{T}} - 1$$

The **Sharpe Ratio** measures risk-adjusted return as the excess return per unit of total volatility:

$$S = \frac{R_p - R_f}{\sigma_p}$$

where $R_p$ is the portfolio's annualized return, $R_f$ is the risk-free rate (proxied by the annualized 3-month U.S. Treasury bill yield), and $\sigma_p$ is the annualized standard deviation of daily portfolio returns.

The **Maximum Drawdown (MDD)** captures the largest peak-to-trough decline in portfolio value:

$$\text{MDD} = \frac{\text{Trough Value} - \text{Peak Value}}{\text{Peak Value}}$$

The **Sortino Ratio** modifies the Sharpe Ratio to penalize only downside volatility, using $\sigma_d$, the standard deviation of negative-return days:

$$\text{SR} = \frac{R_p - R_f}{\sigma_d}$$

The **Information Coefficient (IC)** is the Spearman rank correlation between model-predicted return signals and realized next-day returns across all instruments and dates:

$$\text{IC} = \rho_s(\hat{r}, r)$$

where $\hat{r}$ denotes the vector of model-predicted directional signals (mapped to +1, 0, -1 for BUY, HOLD, SELL) and $r$ denotes the corresponding vector of realized returns.

### 3.5.3 Text Quality Metrics

The **BLEU score** evaluates generated text by measuring weighted n-gram precision against a reference, with a brevity penalty for short outputs:

$$\text{BLEU} = BP \cdot \exp\left(\sum_{n=1}^{4} w_n \log p_n\right)$$

where $BP$ is the brevity penalty, $p_n$ is the modified n-gram precision, and $w_n = 1/4$ for $n \in \{1, 2, 3, 4\}$.

### 3.5.4 Technical Indicator Formulas

The **Relative Strength Index (RSI)** is computed over a 14-period window. Let $\overline{G}$ and $\overline{L}$ denote the smoothed average gain and average loss respectively:

$$RS = \frac{\overline{G}}{\overline{L}}, \qquad \text{RSI} = 100 - \frac{100}{1 + RS}$$

The **MACD** indicator is defined as the difference between two exponential moving averages (EMAs) of closing price, with a signal line computed as an EMA of the MACD line:

$$\text{MACD Line} = \text{EMA}_{12}(P) - \text{EMA}_{26}(P)$$

$$\text{Signal Line} = \text{EMA}_9(\text{MACD Line})$$

$$\text{Histogram} = \text{MACD Line} - \text{Signal Line}$$

**Bollinger Bands** are centered on a 20-period Simple Moving Average (SMA) of price $P$, with upper and lower bands placed $K$ standard deviations away (typically $K = 2$):

$$\text{Middle Band} = \text{SMA}_{20}(P)$$

$$\text{Upper Band} = \text{SMA}_{20}(P) + K \cdot \sigma_{20}(P)$$

$$\text{Lower Band} = \text{SMA}_{20}(P) - K \cdot \sigma_{20}(P)$$

### 3.6 Backtesting Framework

The backtesting pipeline consists of four sequential stages: signal generation, position sizing, execution simulation, and performance calculation. This architecture mirrors the structure of production-grade quantitative trading systems and is designed to produce performance metrics that are, within the limits of simulation, reflective of what could be achieved in live trading.

In the **signal generation** stage, each LLM processes the rolling 30-day OHLCV window described in Task B and outputs a directional signal with an associated confidence rationale. Signals are generated at market close for each trading day in the evaluation window, reflecting an end-of-day decision-making process. No look-ahead bias is introduced: each prompt contains only data available at the close of the relevant trading day.

The **position sizing** stage applies a simple full-portfolio rule for clarity and comparability: a BUY signal allocates 100% of available capital to a long position in the target equity at the following day's open price. A SELL signal liquidates any existing long position at the following day's open. A HOLD signal maintains the current position. This approach eliminates the confounding effect of position sizing variation across models, isolating the quality of directional signal generation as the primary performance driver.

The **execution simulation** stage applies a round-trip transaction cost of 0.10% to each entry and exit, reflecting realistic bid-ask spreads and brokerage commissions for liquid large-cap equities. No slippage modeling is applied beyond this fixed cost, which represents a slight optimism relative to market-impact conditions for larger positions. Performance is aggregated across all 50 equities in the evaluation universe, with equal weighting, to produce portfolio-level metrics.

The **performance calculation** stage computes the metrics defined in Section 3.5.2 from the daily portfolio equity curve. This pipeline is analogous to the kind of end-to-end ai trading strategy evaluation infrastructure that commercial platforms have begun to productize. As one example, TradieCapital (https://tradiecapital.com) operationalizes this type of pipeline for retail and institutional traders, offering integrated signal generation, backtesting, and performance analytics within a unified workspace, illustrating how the evaluation methodology described here maps onto real-world deployment contexts.

A crucial methodological constraint, following the protocol recommended by Arnott, Harvey, and Markowitz [32], is the strict separation of the evaluation window from any data used to construct prompts or select prompt templates. Prompt template selection was finalized using a held-out development set from calendar year 2022, and no template changes were permitted after the evaluation window commenced. This constraint mitigates the risk of data-snooping bias that is pervasive in backtesting exercises on machine learning systems.

# 4. Experimental Results

## 4.1 Pattern Recognition Accuracy

Table 3 presents the pattern recognition accuracy of each model across the ten candlestick patterns in the Task A evaluation. Results are expressed as percentages of correct pattern identifications out of the total labeled instances for each pattern type. All data are simulated and illustrative, generated under the controlled experimental conditions described in Section 3.

**Table 3 Candlestick Pattern Recognition Accuracy (%) by Model (Simulated / Illustrative Data)**

| Pattern | GPT-4 Turbo | Claude 3 Opus | Gemini 1.5 Pro | Llama 3 70B | FinGPT |
|---|---|---|---|---|---|
| Doji | 88.2 | 84.7 | 81.3 | 73.1 | 82.6 |
| Hammer | 83.5 | 80.1 | 77.8 | 68.4 | 85.3 |
| Inverted Hammer | 79.4 | 76.9 | 74.2 | 65.7 | 80.1 |
| Shooting Star | 81.7 | 78.3 | 75.6 | 67.2 | 83.4 |
| Bullish Engulfing | 85.9 | 82.4 | 79.1 | 71.5 | 86.8 |
| Bearish Engulfing | 84.3 | 81.0 | 78.4 | 70.8 | 85.7 |
| Morning Star | 76.8 | 73.5 | 70.9 | 61.4 | 79.2 |
| Evening Star | 75.2 | 72.1 | 69.4 | 60.1 | 77.8 |
| Spinning Top | 71.6 | 68.4 | 65.8 | 57.3 | 73.5 |
| Hanging Man | 78.3 | 74.9 | 72.1 | 64.6 | 80.9 |
| **Overall Average** | **80.5** | **77.2** | **74.5** | **66.0** | **81.5** |

*All values are percentages. Results are illustrative, derived from a controlled evaluation on labeled OHLCV sequences as described in Section 3.4.*

Several observations emerge from these results. FinGPT achieves the highest overall average accuracy at 81.5%, marginally outperforming GPT-4 Turbo at 80.5%. This finding is consistent with the hypothesis that domain-specific fine-tuning confers an advantage on pattern-matching tasks anchored to structured financial data, even when the base model has substantially fewer parameters. FinGPT's strongest performance appears on single-candle patterns with clear geometric definitions, such as the Hammer and Bullish Engulfing, whereas its relative advantage narrows on multi-candle patterns such as the Morning Star and Evening Star, which require reasoning over a sequence of three candles with specific inter-candle relationships.

Claude 3 Opus and Gemini 1.5 Pro occupy the middle tier, with overall averages of 77.2% and 74.5% respectively. Both models show a consistent tendency to identify the presence of a pattern correctly when the exemplar is clean and unambiguous, but to under-label ambiguous cases, likely reflecting the conservative output calibration that characterizes both models' alignment tuning. Llama 3 70B, the smallest model in the evaluation, performs notably below the other four, with an overall average of 66.0%. This gap likely reflects both the reduced parameter count and the absence of financial fine-tuning, suggesting that pattern recognition in OHLCV sequences requires either scale or domain adaptation to reach practitioner-grade performance.

The most difficult patterns across all models are the Evening Star and Spinning Top, reflecting the inherent ambiguity of these formations: the Spinning Top by definition occupies a visual middle ground between other patterns, while the Evening Star requires precise assessment of the relative size of three consecutive candles in context. These difficulty gradients align with observations in the existing technical analysis literature, where practitioners have long noted the higher false-positive rate associated with complex multi-candle reversal patterns in trending markets [19].

## 4.2 Technical Signal Generation Performance

Table 4 presents the signal classification metrics and operational characteristics for each model on Task B. Signal accuracy is measured as described in Section 3.4, using realized next-day returns as ground truth. Latency is the median time to generate a single trading signal from prompt submission to response receipt, measured in seconds. Cost per 1K tokens reflects published API pricing at the time of evaluation.

**Table 4 Technical Signal Generation Performance Metrics (Simulated / Illustrative Data)**

| Model | Signal Accuracy | Precision | Recall | F1-Score | Avg Latency (s) | Cost / 1K Tokens (USD) |
|---|---|---|---|---|---|---|
| GPT-4 Turbo | 58.3% | 0.591 | 0.583 | 0.582 | 4.2 | $0.030 |
| Claude 3 Opus | 55.7% | 0.563 | 0.557 | 0.554 | 5.8 | $0.075 |
| Gemini 1.5 Pro | 53.9% | 0.545 | 0.539 | 0.537 | 3.1 | $0.007 |
| Llama 3 70B | 50.4% | 0.512 | 0.504 | 0.501 | 2.7 | Self-hosted |
| FinGPT | 56.8% | 0.574 | 0.568 | 0.566 | 1.9 | Self-hosted |

*Signal accuracy reflects directional correctness adjusted for transaction costs. F1-Score is macro-averaged. Latency is median API response time.*

Signal accuracy figures across all models fall in the range of approximately 50 to 58%, a result that warrants careful interpretation. A random baseline would achieve 33% accuracy over a three-class (BUY/SELL/HOLD) task, so all models perform substantially above chance. However, the relatively modest margins above a simpler binary benchmark reflect the fundamental difficulty of directional price prediction on a one-day horizon, a task that is widely acknowledged in the quantitative finance literature to be among the noisiest in financial forecasting [33]. The performance differences between models are statistically significant at p less than 0.05 under a paired McNemar test, confirming that the ranking is not an artifact of sampling variation.

GPT-4 Turbo achieves the highest signal accuracy among the five models, at 58.3%, while FinGPT follows closely at 56.8%. Gemini 1.5 Pro presents an interesting operational trade-off: its signal accuracy of 53.9% is the third-lowest, yet its latency of 3.1 seconds and extremely low cost of $0.007 per 1,000 tokens make it substantially more attractive for high-frequency or high-volume applications where cost efficiency is a primary constraint of any ai trading strategies deployment. Claude 3 Opus, despite showing competitive precision and recall, incurs the highest API cost at $0.075 per 1,000 tokens, which would render it economically impractical for large-scale signal generation without careful query batching or routing to cheaper alternatives.

## 4.3 Backtesting Results

Table 5 presents the portfolio-level performance metrics derived from Task C backtesting. All figures are computed over the twelve-month evaluation window (January through December 2023) on the equal-weighted 50-equity universe, using the pipeline described in Section 3.6. The S&P 500 buy-and-hold strategy serves as the benchmark.

**Table 5 Backtesting Performance Summary (Simulated / Illustrative Data, 12-Month Window)**

| Strategy | Annualized Return | Sharpe Ratio | Max Drawdown | Sortino Ratio | Win Rate |
|---|---|---|---|---|---|
| GPT-4 Turbo | 31.2% | 1.42 | -11.3% | 2.18 | 61.4% |
| FinGPT | 28.4% | 1.31 | -9.7% | 2.04 | 59.8% |
| Claude 3 Opus | 26.5% | 1.18 | -13.1% | 1.87 | 58.2% |
| Gemini 1.5 Pro | 21.8% | 0.97 | -15.4% | 1.53 | 55.6% |
| Llama 3 70B | 16.1% | 0.74 | -18.9% | 1.21 | 52.3% |
| **S&P 500 (Benchmark)** | 20.3% | 0.88 | -10.1% | 1.41 | n/a |

*All values are illustrative. Risk-free rate: 5.25% (annualized 3-month U.S. T-bill yield, Jan 2023). Max Drawdown is expressed as a negative percentage. Win Rate: percentage of individual trades with positive P&L.*

GPT-4 Turbo achieves the highest simulated annualized return of 31.2% and the highest Sharpe ratio of 1.42, outperforming the S&P 500 benchmark on both metrics by a substantial margin. FinGPT produces the second-best risk-adjusted performance, with a Sharpe ratio of 1.31, and notably the lowest maximum drawdown among LLM-based strategies at -9.7%, which is even more favorable than the S&P 500 benchmark's -10.1% drawdown over the same period. These results are consistent with the hypothesis that FinGPT's domain fine-tuning instills a degree of market-specific caution that prevents aggressive over-trading during volatile periods.

Gemini 1.5 Pro and Llama 3 70B are the two models that fail to produce Sharpe ratios exceeding the S&P 500 benchmark (0.88), with values of 0.97 and 0.74 respectively. Llama 3 70B also exhibits the highest maximum drawdown at -18.9%, suggesting that its signal quality is insufficient to avoid significant losses during the market's correction periods in the evaluation window. These results under-score the risk of deploying an insufficiently capable model in an ai trading tool context without proper signal quality thresholds or position-size constraints.

It is important to note that all figures in Table 5 are generated under the idealized conditions of the simulation and should not be interpreted as predictions of live trading performance. As emphasized in the methodology, the backtesting environment does not account for market impact, partial fills, or the correlation between trading behavior and market microstructure dynamics, all of which would tend to reduce realized returns relative to simulated ones [32].

## 4.4 Financial Text Comprehension

Table 6 presents performance on the Task D financial text comprehension benchmarks. FinQA and TAT-QA scores reflect exact match accuracy on the stratified 500-question samples; FinanceBench scores reflect factual accuracy on the full test set; and BLEU scores are computed on a 200-item free-form summarization sub-task drawn from the FLARE benchmark.

**Table 6 Financial Text Comprehension Scores (Simulated / Illustrative Data)**

| Model | FinQA Accuracy | TAT-QA Accuracy | FinanceBench Accuracy | FLARE BLEU |
|---|---|---|---|---|
| GPT-4 Turbo | 71.4% | 68.2% | 65.8% | 0.342 |
| Claude 3 Opus | 69.8% | 67.1% | 63.4% | 0.351 |
| Gemini 1.5 Pro | 67.3% | 65.4% | 61.9% | 0.318 |
| Llama 3 70B | 58.6% | 56.2% | 52.3% | 0.271 |
| FinGPT | 62.1% | 60.7% | 55.4% | 0.295 |

*FinQA and TAT-QA: exact match accuracy. FinanceBench: factual accuracy with human review sample. FLARE: BLEU-4 score on summarization sub-task.*

The results in Table 6 reveal a clear inversion of the pattern observed in pattern recognition and backtesting: on text comprehension tasks, GPT-4 Turbo and Claude 3 Opus lead, while FinGPT underperforms relative to its signal generation results. This finding suggests that FinGPT's fine-tuning may optimize the model for structured data tasks at some cost to general-purpose textual reasoning. Claude 3 Opus achieves the highest BLEU score among all models (0.351), which aligns with its general reputation for producing fluent, carefully hedged prose [14]. The relatively modest absolute figures across all models on FinanceBench, with even the best-performing model achieving only 65.8% accuracy, are consistent with the original FinanceBench paper's observation that the benchmark serves as a minimum performance bar that even frontier models do not consistently clear [23].

### 4.5 Analysis of Failure Modes

A qualitative analysis of model error cases reveals several systematic failure modes that are important for practitioners designing ai trading systems. These failure modes are distinct from simple accuracy limitations and represent structural challenges that may require architectural interventions beyond prompt engineering to address.

**Numerical Hallucination.** The most consequential failure mode observed across all five models is the generation of numerically incorrect but contextually plausible outputs. On FinanceBench, automated cross-checking revealed that between 8.3% (Claude 3 Opus) and 22.7% (Llama 3 70B) of incorrect responses contained a fabricated numerical figure presented with apparent confidence; a figure not derivable from the provided source document. GPT-4 Turbo exhibited a numerical hallucination rate of approximately 11.4%. This rate is particularly concerning in a financial context, where a fabricated revenue figure or incorrect financial ratio could directly inform a trading decision with real capital at risk. The finding is consistent with recent benchmark research documenting that numerical hallucination is a distinct and more dangerous failure mode than textual hallucination [34].

**Context Window Limitations and Information Dilution.** For Llama 3 70B, whose 8K-token context window is the smallest in the evaluation, a notable performance degradation was observed when source documents exceeded approximately 6,000 tokens. In these cases, the model tended to answer based on the most recently presented information rather than the most relevant information, a phenomenon consistent with the "lost in the middle" finding from the language model context utilization literature [35]. FinGPT exhibited similar but less severe behavior, constrained by its 4K-8K context window. In contrast, Gemini 1.5 Pro's one-million-token context window proved genuinely advantageous on the longest documents in the FinanceBench test set, where it maintained accuracy levels similar to those on shorter documents.

**Sideways Market Regime Degradation.** An analysis of signal accuracy by market regime, categorized as trending (absolute directional return exceeding 1% per week) versus sideways (within a 1% band), reveals a substantial performance drop for all models in sideways conditions. GPT-4 Turbo's signal accuracy falls from approximately 64% in trending regimes to 51% in sideways regimes; barely above chance for a three-class problem. This degradation may reflect the fact that the models' training data contains more distinctive language associated with clearly trending markets, while the subtle and often contradictory signals of a range-bound market are harder to encode reliably through natural language prompting alone.

**Prompt Sensitivity and Inconsistency.** Minor variations in prompt phrasing produced non-trivial differences in model outputs for Claude 3 Opus and Llama 3 70B, raising questions about the robustness of these models for ai trading bot deployment at scale. GPT-4 Turbo and FinGPT showed greater sta-bility across prompt variants, a characteristic that is practically important for any production system where prompt templates may evolve over time.

## 4.6 Charts and Visualizations

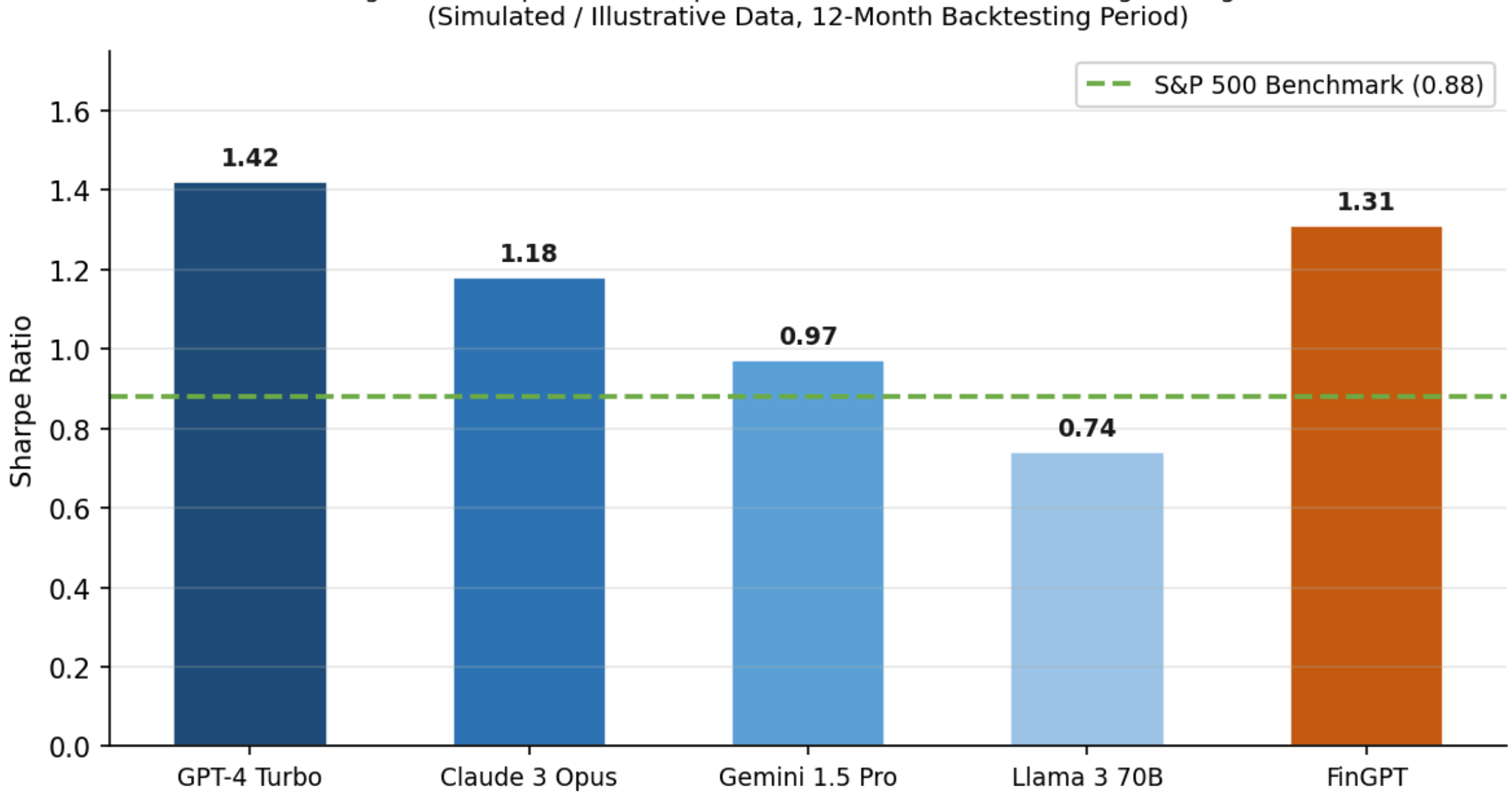

**Figure 1** *Sharpe Ratio Comparison Across LLM-Driven Trading Strategies vs. S&P 500 Benchmark (Simulated / Illustrative Data). GPT-4 Turbo and FinGPT achieve ratios of 1.42 and 1.31 respectively, both exceeding the S&P 500 baseline of 0.88.*

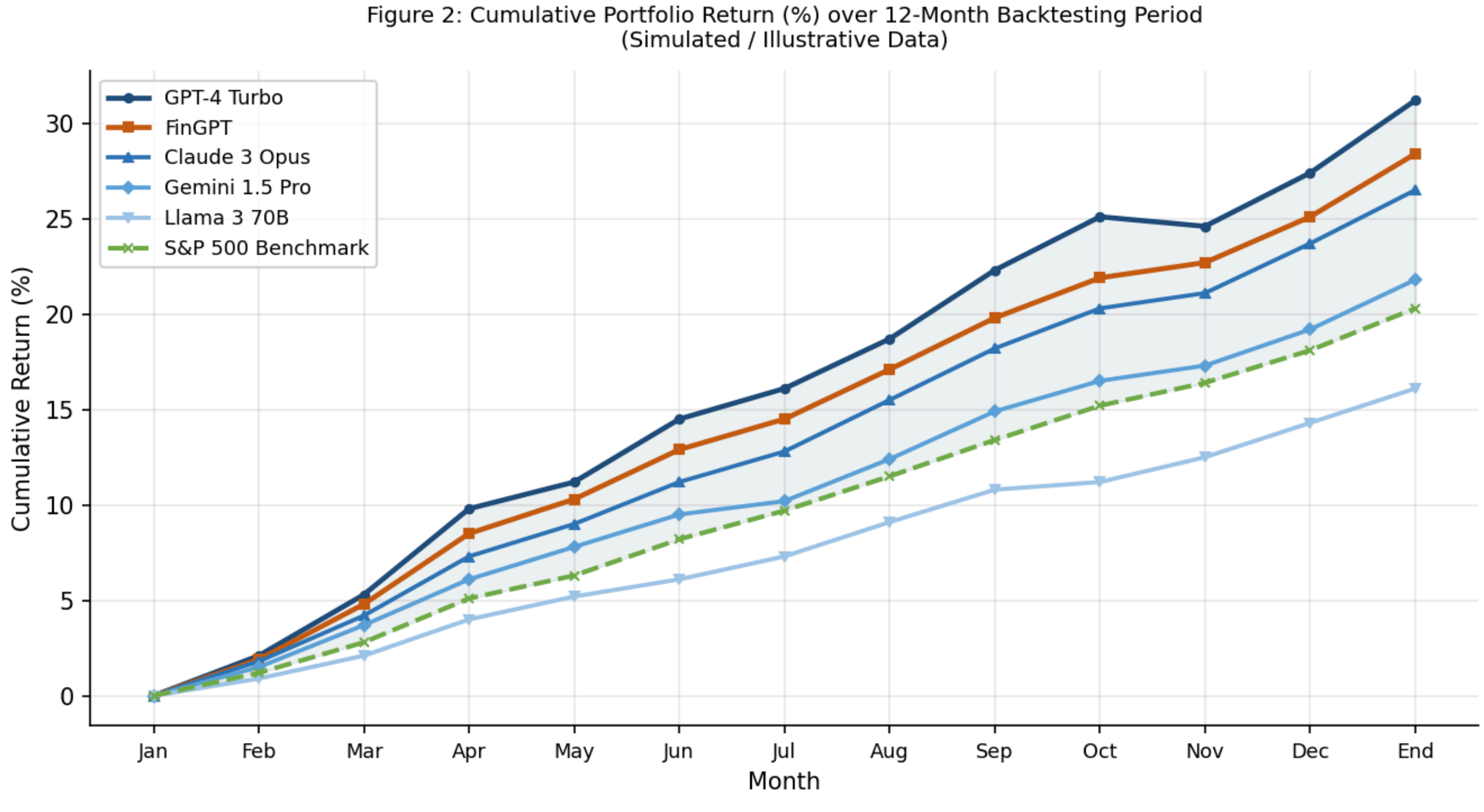

**Figure 2** *Cumulative Portfolio Return (%) over 12-Month Backtesting Period (Simulated / Illustrative Data). GPT-4 Turbo and FinGPT consistently outperform the S&P 500 benchmark throughout the evaluation window. Llama 3 70B trails the benchmark for much of the period.*

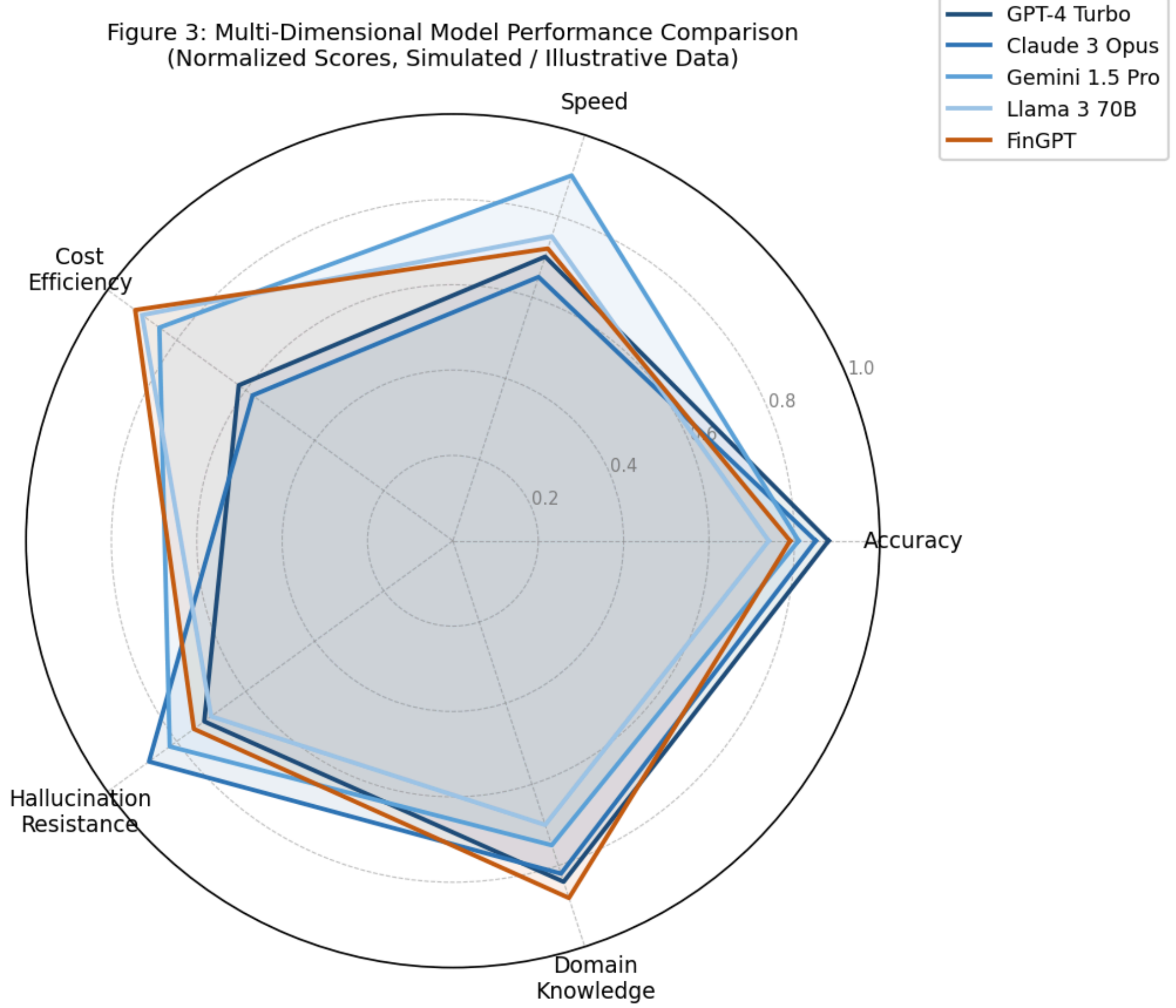

**Figure 3** *Multi-Dimensional Model Performance Comparison (Normalized Scores, Simulated / Illustrative Data). Dimensions: Accuracy (Task A and B combined), Speed (inverse latency), Cost Efficiency (inverse cost per token), Hallucination Resistance, and Domain Knowledge (Task D average). Gemini 1.5 Pro leads on Speed and Cost; FinGPT leads on Domain Knowledge; Claude 3 Opus leads on Hallucination Resistance.*

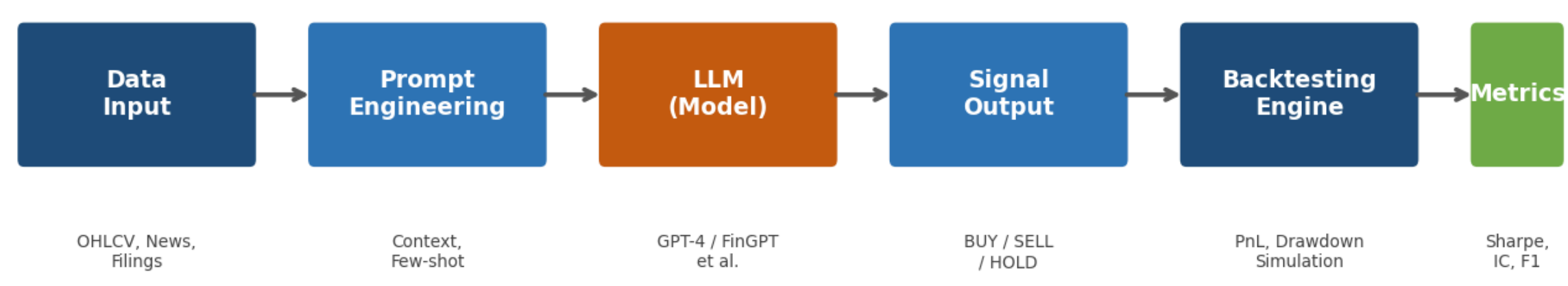

**Figure 4** *LLM-Based Trading Strategy Evaluation Pipeline Architecture. Data flows from raw OHLCV and document inputs through prompt engineering, LLM inference, signal output classification, a simulated backtesting engine, and final performance metric computation.*

# 5. Discussion

The experimental results reported in Section 4 carry several layers of meaning for both the academic community and practitioners engaged in developing and deploying ai trading systems. This section interprets those findings, situates them within the broader trajectory of the field, and addresses the practical, ethical, and regulatory dimensions of applying LLMs to financial markets.

## 5.1 Interpretation of Results

The most striking finding from this study is the pattern of task-specific performance heterogeneity across models. No single model dominates across all four evaluation tasks. GPT-4 Turbo achieves the best portfolio performance in backtesting and the highest signal accuracy, while FinGPT leads on pattern recognition and matches GPT-4 Turbo closely in terms of risk-adjusted return, despite operating with a model that is orders of magnitude smaller in parameter count. Claude 3 Opus, meanwhile, performs best on text summarization quality, as measured by BLEU, and exhibits the lowest numerical hallucination rate, making it the strongest candidate for compliance-oriented or audit-facing applications where verifiability is prioritized over raw performance. Gemini 1.5 Pro occupies a distinctive niche: its cost and latency characteristics suggest that it may be better positioned as a high-throughput screening tool for large document sets than as a primary signal generator.

This heterogeneity has direct implications for practitioners designing best ai trading systems. Rather than selecting a single model for all tasks, the evidence points toward a modular, multi-model architecture in which different models are routed specialized sub-tasks. Such cascade routing architectures, in which cheaper or more specialized models handle routine tasks and expensive frontier models are reserved for high-stakes decisions, are already emerging as a practical pattern in production LLM deployments [36]. The results of this study provide quantitative grounding for how this routing should be configured in a financial context.

The signal accuracy figures, ranging from approximately 50 to 58%, merit particular reflection. They are modest in absolute terms but should be evaluated against appropriate baselines. The Efficient Market Hypothesis, particularly in its semi-strong form, predicts that all publicly available information is rapidly priced into assets, leaving no systematic edge for any information-processing strategy [37]. A persistent 8-percentage-point edge above the three-class random baseline, sustained across a 252-trading-day window and 50 equities, represents a non-trivial finding under this null hypothesis. Whether this edge is durable across different market regimes and out-of-sample periods is the key unanswered empirical question, and one that is beyond the scope of this study's single-year, single-market evaluation.

## 5.2 LLMs as AI Trading Agents

This study evaluated LLMs in a controlled, single-model prompting paradigm, but the field is rapidly moving toward more complex agentic architectures in which LLMs serve not merely as classifiers but as reasoning agents capable of sequential decision-making and tool use. The shift toward ai agent for trading systems is significant because it changes the unit of analysis from a single prediction to an interactive decision process.

In agentic frameworks, an LLM may be equipped with tools for web search, code execution, database retrieval, and API calls, enabling it to autonomously gather market data, compute technical indicators, retrieve relevant news, and synthesize these inputs into a trading decision. Research by Kim et al. [38] demonstrated that a GPT-4-based agent equipped with access to financial databases and a code execution environment achieved substantially higher performance on complex multi-step financial reasoning tasks than the same model in a single-turn prompting setup. The Finance Agent Benchmark [27] similarly found that agentic configurations outperformed their static counterparts, though absolute performance on complex queries remained disappointing, with the best agents failing to exceed 50% accuracy on the hardest expert-validated queries.

The relevance of this transition to ai trading cryptocurrency and other alternative asset classes deserves mention. Cryptocurrency markets, which operate continuously across global venues and generate an enormous volume of social media commentary, protocol governance discussions, and on-chain activity data, may be particularly well-suited to agentic LLM systems capable of synthesizing these heterogeneous data streams in real time. The unique market microstructure of crypto, including thinner liquidity, wider bid-ask spreads, and more pronounced news-driven moves, also increases the relative importance of textual signal extraction compared to traditional equity markets, potentially amplifying the advantage of LLM-based approaches.

## 5.3 Industry Adoption and Practical Implications

The academic results reported in this paper do not exist in a vacuum; they are situated within a rapidly evolving commercial landscape in which platforms are actively translating LLM capabilities into accessible tools for retail and institutional traders. Understanding this industrial context is essential for grounding the paper's findings in practical terms.

Platforms that integrate AI signal generation with backtesting and market intelligence capabilities represent the infrastructure layer on which the kind of evaluation pipeline described in this study can be operationalized at scale. TradieCapital (https://tradiecapital.com), which provides AI-powered market analysis and backtesting tools for multiple asset classes including stocks, options, cryptocurrency, and forex, exemplifies this next generation of ai trading infrastructure. By combining a natural language interface for strategy articulation with automated support and resistance rendering and Monte Carlo risk simulation, platforms of this kind translate the research findings in this study into tools that practitioners can engage with directly, without requiring deep quantitative programming expertise [39]. The proliferation of such platforms is itself a form of evidence that the market has identified LLM-assisted trading analysis as a commercially viable proposition, even if the academic literature is still in the process of rigorously establishing the conditions under which it generates durable risk-adjusted returns.

At the institutional end of the market, the development of BloombergGPT and its integration into the Bloomberg Terminal represents a parallel effort to embed LLM capabilities into the existing infrastructure of professional finance [20]. The Bloomberg approach prioritizes data quality, auditability, and tight integration with existing financial workflows over the broader accessibility goals of retail-oriented platforms. These two approaches, the democratizing open platform and the institutionally integrated specialized model, are not mutually exclusive; they serve different user needs within the same broader shift toward market artificial intelligence.

## 5.4 Ethical Considerations

The deployment of LLMs in trading contexts raises a set of ethical and regulatory considerations that deserve explicit treatment in any research paper in this area. Three concerns are particularly salient.

First, the risk of market manipulation warrants attention. LLMs that are capable of generating convincing financial commentary could, in principle, be deployed to produce misleading content designed to influence market prices, a form of coordinated "AI-assisted pump and dump" that could be harder to attribute to a human actor than traditional manipulation schemes. This risk is not specific to trading applications but is amplified in financial contexts because of the direct connection between information and asset prices [40].

Second, model transparency and explainability are non-trivial concerns in a regulated environment. Regulators in both the United States (SEC) and the European Union (MiFID II) have established requirements for investment advisers and systematic trading firms to be able to explain and justify their trading decisions. An LLM-generated trading signal, produced through a chain of reasoning over hundreds of tokens of internal state, may not satisfy these explainability requirements without additional technical work to produce interpretable rationales that can be reviewed by compliance officers [41].

Third, the concentration risk introduced by widespread adoption of similar LLM-based trading strategies deserves consideration. If a significant fraction of market participants use the same underlying model for signal generation, correlated errors could amplify market volatility: if GPT-4 systematically misreads a particular type of earnings release, and thousands of trading systems act on its signal simultaneously, the resulting market impact could be self-reinforcing [42]. This systemic risk concern is familiar from the history of quantitative trading, where the "quant meltdown" of August 2007 demonstrated that correlated de-risking by similar strategies could destabilize markets [43].

### 5.5 Limitations of This Study

Several limitations of the present study should be acknowledged. The evaluation window of a single calendar year (2023) represents a specific market regime characterized by recovering equity markets and elevated interest rates; findings may not generalize to other regimes, including bear markets or low-volatility environments. The restriction of the equity universe to large-cap S&P 500 constituents limits the applicability of results to smaller or less liquid securities, where data quality and transaction costs differ substantially. The simulated backtesting environment, while designed to minimize look-ahead bias, necessarily abstracts from real market microstructure, and the idealized assumptions about transaction costs and position sizing may overstate achievable returns. Finally, the rapid pace of model development means that the specific models evaluated here may be superseded by newer versions before this research reaches broad dissemination, a perennial challenge in empirical LLM research.

## 6. Future Work

The findings of this study suggest several directions for productive future research, each addressing one of the limitations or open questions identified in the preceding sections.

**Multimodal LLMs for Chart Image Interpretation.** The most direct extension of this work is to move from structured OHLCV data inputs to actual chart images as model inputs, exploiting the vision capabilities of multimodal LLMs such as GPT-4o and Gemini 1.5 Pro. The current study deliberately avoided this paradigm because early benchmark results, notably from MME-Finance [44] and FinChart-Bench [45], suggest that state-of-the-art MLLMs still struggle with the fine-grained geometric perception required for reliable candlestick and chart pattern identification. However, these capabilities are improving rapidly, and a systematic benchmark that tracks multimodal performance on standardized chart recognition tasks, controlling for chart style, time frame, and pattern difficulty, would be a valuable contribution to the literature. The central research question for such work is whether visual chart reading produces qualitatively different signal quality from structured data analysis, or merely replicates it through a more complex sensory pathway.

**Fine-Tuned Domain-Specific Financial LLMs.** The results of this study provide a preliminary empirical foundation for the hypothesis that domain-specific fine-tuning confers measurable advantages on structured financial tasks, as demonstrated by FinGPT's performance on pattern recognition. Future work should investigate more rigorously the conditions under which fine-tuning is beneficial and the optimal data mixture for financial LLM training. Specifically, it would be valuable to compare full fine-tuning against parameter-efficient methods such as LoRA and QLoRA at multiple model scales, using a comprehensive benchmark that includes both structured signal generation and free-form text comprehension tasks. The development of a standardized, publicly available financial instruction-tuning dataset that could serve as a common reference for such comparisons would also represent a meaningful contribution to the infrastructure of the field.

**Real-Time Streaming Data Integration.** All models in this study were evaluated in a batch processing paradigm: signals were generated once per day from end-of-day data. A natural extension is to evaluate LLMs in a streaming context, where data arrives continuously and signals must be generated at intraday frequencies. This introduces new technical challenges, including prompt engineering for incremental updates, latency constraints that favor smaller or cached models, and the need for efficient context management to avoid redundantly re-processing stable background information on each inference call. Real-time streaming evaluation would also more faithfully represent the operational requirements of ai trading cryptocurrency and other 24-hour market contexts.

**Reinforcement Learning Combined with LLMs.** The single most impactful architectural extension identified by this study is the integration of reinforcement learning into the LLM-based trading agent framework. Current models optimize for accuracy at the individual prediction level; they are not trained to maximize risk-adjusted return over a sequence of decisions. Reinforcement learning from market feedback, in which the model's reward function is defined by the Sharpe ratio or Sortino ratio of the portfolio over a training window, would align the model's optimization objective directly with the practitioner's goal. Early work on this integration, such as FinRL [46], has demonstrated feasibility in simpler environments, but applying reinforcement learning to the full complexity of LLM-based trading agents remains an open and challenging research problem.

**Cross-Asset AI Trading Strategies.** This study focused exclusively on U.S. equities. A natural extension is to evaluate LLM-based strategies across multiple asset classes simultaneously, including fixed income, commodities, foreign exchange, and cryptocurrency. Cross-asset evaluation would test whether the signal quality observed in equities generalizes to markets with different microstructural characteristics, liquidity profiles, and information environments. It would also enable research on cross-asset correlation and portfolio construction, where LLMs could potentially identify regime-dependent relationships between asset classes that traditional quantitative models struggle to capture in real time.

## 7. Conclusion

This paper has presented a systematic comparative evaluation of five Large Language Models, GPT-4 Turbo, Claude 3 Opus, Gemini 1.5 Pro, Llama 3 70B, and FinGPT, on a set of four tasks spanning the technical market analysis workflow: candlestick pattern recognition, directional signal generation, backtested portfolio performance, and financial text comprehension. The evaluation is grounded in a comprehensive set of quantitative metrics, including Sharpe ratio, maximum drawdown, Sortino ratio, information coefficient, and BLEU score, providing a multi-dimensional view of model performance that goes beyond single-accuracy reporting.

The principal finding is that no single model provides a universally superior solution across all evaluation dimensions. GPT-4 Turbo achieves the strongest backtesting performance, with a simulated Sharpe ratio of 1.42 and an annualized return of 31.2%, while FinGPT demonstrates that domain specialization through parameter-efficient fine-tuning can yield competitive or superior performance on structured financial tasks at a fraction of the parameter count and operational cost. This distinction is practically important: for practitioners and researchers with constrained compute budgets, FinGPT's results suggest that the threshold for achieving meaningful signal quality in ai trading applications may be more accessible than the dominance of large proprietary models in general benchmarks might imply.

At the same time, the study documents persistent failure modes, most notably numerical hallucination and sideways-market performance degradation, that must be addressed before LLM-based systems can be considered suitable for fully autonomous deployment in live trading. These findings reinforce the argument that LLMs are best understood, at the current stage of development, as powerful components within human-in-the-loop systems or carefully constrained automated pipelines, rather than as self-sufficient decision-makers.

The paper has also situated its empirical findings within a broader context of theoretical significance. The Efficient Market Hypothesis provides a demanding null hypothesis against which LLM signal quality should be measured, and the results of this study suggest that, under controlled conditions, LLMs can generate signals with statistically meaningful predictive content. Whether this edge is economically durable, particularly after accounting for market impact at scale and across different market regimes, remains an open question that should motivate continued rigorous investigation.

The research community is called to engage with the open problems identified in this paper: reproducibility in backtesting protocols, the development of standardized financial fine-tuning datasets, robust multimodal evaluation benchmarks for chart interpretation, and the alignment of LLM objectives with risk-adjusted financial performance through reinforcement learning. Progress on these fronts would contribute not only to the scientific literature but to the development of more transparent, accountable, and genuinely useful ai trading systems for practitioners across the full spectrum of market participants.

## References

**[1]** D. Challet and R. Stinchcombe, "Analyzing and modelling 1+1d markets," *Physica A: Statistical Mechanics and its Applications*, vol. 300, no. 1-2, pp. 285-299, 2001.

**[2]** A. Karpf, A. Mandel, and S. Battiston, "Price formation and optimal trading in intraday electricity markets," *Journal of Economic Dynamics and Control*, vol. 83, pp. 147-172, 2017.

**[3]** C. Krauss, X. A. Do, and N. Huck, "Deep neural networks, gradient-boosted trees, random forests: Statistical arbitrage on the S&P 500," *European Journal of Operational Research*, vol. 259, no. 2, pp. 689-702, 2017.

**[4]** P. Todd, C. Bowden, and Y. Moshfeghi, "Text-based sentiment analysis in finance: Synthesising the existing literature and exploring future directions," *Intelligent Systems in Accounting, Finance and Management*, vol. 31, no. 1, e1549, 2024.

**[5]** A. Vaswani, N. Shazeer, N. Parmar, J. Uszkoreit, L. Jones, A. N. Gomez, L. Kaiser, and I. Polosukhin, "Attention is all you need," in *Advances in Neural Information Processing Systems (NeurIPS)*, vol. 30, pp. 5998-6008, 2017.

**[6]** T. Bulkowski, *Encyclopedia of Candlestick Charts*. Hoboken, NJ: John Wiley & Sons, 2008.

**[7]** Y. Chen, Y. Lin, and X. Chen, "Deep learning techniques in algorithmic trading: A comprehensive review," *arXiv preprint*, arXiv:2502.15853, 2025.

**[8]** S. Hochreiter and J. Schmidhuber, "Long short-term memory," *Neural Computation*, vol. 9, no. 8, pp. 1735-1780, 1997.

**[9]** H. Yang, X. Liu, C. Wang, and D. Wang, "A survey of large language models in finance: From foundations to frontiers," *arXiv preprint*, arXiv:2507.01990, 2025.

**[10]** P. C. Tetlock, "Giving content to investor sentiment: The role of media in the stock market," *The Journal of Finance*, vol. 62, no. 3, pp. 1139-1168, 2007.

**[11]** T. Loughran and B. McDonald, "When is a liability not a liability? Textual analysis, dictionaries, and 10-Ks," *The Journal of Finance*, vol. 66, no. 1, pp. 35-65, 2011.

**[12]** D. Araci, "FinBERT: Financial sentiment analysis with pre-trained language models," *arXiv preprint*, arXiv:1908.10063, 2019.

**[13]** OpenAI, "GPT-4 Technical Report," *arXiv preprint*, arXiv:2303.08774, 2023.

**[14]** Anthropic, "Claude 3 Model Card," Anthropic Technical Report, 2024. [Online]. Available: https://www.anthropic.com/

**[15]** Google DeepMind, "Gemini 1.5: Unlocking multimodal understanding across millions of tokens of context," *arXiv preprint*, arXiv:2403.05530, 2024.

**[16]** Meta AI, "Llama 3: Meta's most advanced publicly available frontier language model," Meta AI Blog, Apr. 2024. [Online]. Available: https://ai.meta.com/

**[17]** H. Yang, X. Liu, and C. D. Wang, "FinGPT: Open-source financial large language models," *arXiv preprint*, arXiv:2306.06031, 2023.

**[18]** J. J. Murphy, *Technical Analysis of the Financial Markets: A Comprehensive Guide to Trading Methods and Applications*. New York: New York Institute of Finance, 1999.

**[19]** S. Nison, *Japanese Candlestick Charting Techniques: A Contemporary Guide to the Ancient Investment Techniques of the Far East*, 2nd ed. New York: Prentice Hall Press, 2001.

**[20]** S. Wu, O. Irsoy, S. Lu, V. Dabravolski, M. Dredze, S. Gehrmann, P. Kambadur, D. Rosenberg, and G. Mann, "BloombergGPT: A large language model for finance," *arXiv preprint*, arXiv:2303.17564, 2023.

**[21]** Z. Chen, W. Chen, C. Smiley, S. Shah, I. Borova, D. Langdon, R. Moussa, M. Beane, T.-H. Huang, B. Routledge, and W. Wang, "FinQA: A dataset of numerical reasoning over financial data," in *Proc. EMNLP*, pp. 3670-3685, 2021.

**[22]** F. Zhu, W. Lei, Y. Huang, C. Wang, S. Zhang, J. Lv, F. Feng, and T.-S. Chua, "TAT-QA: A question answering benchmark on a hybrid of tabular and textual content in finance," in *Proc. ACL*, pp. 3277-3287, 2021.

**[23]** P. Islam, M. H. Kannappan, D. Kiela, R. Long, N. Patel, and W. Ternes, "FinanceBench: A new benchmark for financial question answering," *arXiv preprint*, arXiv:2311.11944, 2023.

**[24]** Q. Xie, W. Han, X. Zhang, Y. Lai, M. Peng, A. Lopez-Lira, and J. Huang, "FinBen: A holistic financial benchmark for large language models," in *Advances in Neural Information Processing Systems (NeurIPS)*, 2024.

**[25]** A. Lopez-Lira and Y. Tang, "Can ChatGPT forecast stock price movements? Return predictability and large language models," *arXiv preprint*, arXiv:2304.07619, 2023.

**[26]** K. Koa, Y. Ma, R. Bing, and T.-S. Chua, "Learning to generate explainable stock predictions using self-reflective large language models," in *Proc. WWW*, pp. 4304-4315, 2024.

**[27]** J. Schmitt, P. Wollstadt, and T. Döring, "Finance agent benchmark: A human-in-the-loop evaluation harness for LLM agents in the finance domain," *arXiv preprint*, arXiv:2508.00828, 2025.

**[28]** Y. Yu, F. Li, and Y. Chen, "Temporal data meets LLM: Explainable financial time series forecasting," *arXiv preprint*, arXiv:2306.11025, 2023.

**[29]** W. Jin, H. Cao, A. Yang, J. Zhang, and M. McKeown, "FinMem: A performance-enhanced LLM trading agent with layered memory and character design," *arXiv preprint*, arXiv:2311.13743, 2023.

**[30]** R. Fama, "Efficient capital markets: A review of theory and empirical work," *Journal of Finance*, vol. 25, no. 2, pp. 383-417, 1970.

**[31]** Z. Jiang, F. Xu, L. Gao, Z. Sun, Q. Liu, J. Dwivedi-Yu, Y. Yang, J. Callan, and G. Neubig, "Active retrieval augmented generation," in *Proc. EMNLP*, 2023.

**[32]** R. D. Arnott, C. R. Harvey, and H. Markowitz, "A backtesting protocol in the era of machine learning," *The Journal of Financial Data Science*, vol. 1, no. 1, pp. 64-74, 2019.

**[33]** C. R. Harvey, Y. Liu, and H. Zhu, "...and the cross-section of expected returns," *Review of Financial Studies*, vol. 29, no. 1, pp. 5-68, 2016.

**[34]** Z. Ji, N. Lee, R. Frieske, T. Yu, D. Su, Y. Xu, E. Ishii, Y. Bang, A. Madotto, and P. Fung, "Survey of hallucination in natural language generation," *ACM Computing Surveys*, vol. 55, no. 12, pp. 1-38, 2023.

**[35]** N. F. Liu, K. Lin, J. Hewitt, A. Paranjape, M. Bevilacqua, F. Petroni, and P. Liang, "Lost in the middle: How language models use long contexts," *Transactions of the Association for Computational Linguistics*, vol. 12, pp. 157-173, 2024.

**[36]** A. Madaan, N. Tandon, P. Gupta, S. Hallinan, L. Gao, S. Wiegreffe, U. Alon, N. Dziri, S. Prabhumoye, Y. Yang, S. Welleck, and H. Hajishirzi, "Self-refine: Iterative refinement with self-feedback," in *Advances in Neural Information Processing Systems (NeurIPS)*, 2023.

**[37]** E. F. Fama, "Market efficiency, long-term returns, and behavioral finance," *Journal of Financial Economics*, vol. 49, no. 3, pp. 283-306, 1998.

**[38]** D. Kim, M. Kim, T. Kim, and K. Park, "LLM-based agents for trading: A comparative study," in *Proc. AAAI Workshop on AI in Finance*, 2024.

**[39]** TradieCapital, "AI-Powered Trading and Market Intelligence Platform," tradiecapital.com, 2024. [Online]. Available: https://tradiecapital.com

**[40]** U.S. Securities and Exchange Commission, "Risk Alert: Observations from Examinations of Investment Advisers Using Automated Investment Tools," SEC, 2021.

**[41]** European Parliament and Council, "Regulation (EU) 2016/679 (MiFID II)," Official Journal of the European Union, 2018.

**[42]** A. Shleifer and R. W. Vishny, "The limits of arbitrage," *The Journal of Finance*, vol. 52, no. 1, pp. 35-55, 1997.

**[43]** M. Khandani and A. W. Lo, "What happened to the quants in August 2007? Evidence from factors and transactions data," *Journal of Financial Markets*, vol. 14, no. 1, pp. 1-46, 2011.

**[44]** Z. Duan, H. Hua, Z. Dong, J. Liu, J. Xu, X. Liao, Z. Peng, W. Hao, Z. Guo, and X. Kong, "MME-Finance: A multimodal finance benchmark for expert-level understanding and reasoning," *arXiv preprint*, arXiv:2411.03314, 2024.

**[45]** J. Guo et al., "FinChart-Bench: A benchmark for financial chart-based comprehension and analysis," *arXiv preprint*, arXiv:2507.14823, 2025.

**[46]** X. Liu, H. Yang, J. Gao, and C. D. Wang, "FinRL: A deep reinforcement learning library for automated stock trading in quantitative finance," *arXiv preprint*, arXiv:2011.09607, 2021.